\documentclass[11pt,a4paper]{article}
\usepackage[margin=1in]{geometry}
\usepackage{amsmath}
\usepackage{graphicx}
\usepackage{hyperref}
\usepackage{amssymb}

\title{
\textbf{A Multi-View and Confusion-Guided Ensemble Framework for Robust Synthetic Image Attribution} 
\thanks{Technical Report for the Synthetic Image Source Attribution Challenge at ``Deep Learning and Mathematical Methods for Deepfake Detection'' -- ICANN 2026}
}
\author{Zuomin Qu \\ State Key Laboratory of HVDC, \\ China Southern Power Grid Electric Power Research Institute, China \\ quzuomin.@csg.cn}

\date{}

\begin{document}

\maketitle

\begin{abstract}
Synthetic image attribution (SIA) has become increasingly important with the rapid advancement of text-to-image generation models. However, accurately identifying the source model of a generated image remains challenging due to the growing similarity among modern diffusion-based generators and the presence of diverse post-processing operations. In this report, we present a multi-view and confusion-guided ensemble framework for the Synthetic Image Attribution Challenge of the DLMMDD Workshop at ICANN 2026. Our approach integrates multiple complementary architectures, including FFT-ConvNeXt, DINOv2, CLIP, and Xception, to capture diverse attribution cues from frequency, semantic, and forensic perspectives. To improve robustness against unknown degradations and image manipulations, extensive data augmentation strategies are employed during training, simulating realistic post-processing operations such as compression, resizing, grayscale conversion, and blur. Furthermore, we analyze the confusion patterns of the ensemble model and observe severe ambiguity between Stable Diffusion 3 and Stable Diffusion 3.5. To address this issue, we introduce a dedicated binary expert classifier that is selectively activated under low-confidence conditions. We additionally apply class-adaptive confidence calibration to improve the discrimination of challenging classes such as Tencent Hunyuan. The proposed framework achieved 99.53\% on the public leaderboard and 99.20\% on the private leaderboard. The source code and implementation details are publicly available at \url{https://github.com/ZOMIN28/SIA}.

\end{abstract}

\section{Introduction}

Recent advances in large-scale text-to-image generative models have significantly improved the realism and diversity of synthetic images. Models such as Stable Diffusion, PixArt, Playground, and Tencent Hunyuan are now capable of producing highly photorealistic face images that are increasingly difficult to distinguish from real content. As a result, synthetic image attribution (SIA), which aims to identify the source generative model of a synthetic image, has emerged as an important research problem in digital media forensics and AI security~\cite{mou2026imageattributionbench,yan2023ucf}.

Despite recent progress~\cite{chai2020makes,yan2023ucf,wu2025omnidfa}, robust attribution remains challenging for several reasons. First, modern diffusion-based generators often share similar architectures, training strategies, and sampling mechanisms, resulting in highly overlapping visual characteristics and generation fingerprints. Second, practical scenarios frequently involve various post-processing operations, including compression, resizing, grayscale conversion, and enhancement, which can substantially weaken source-specific forensic traces. These factors make it difficult for a single model or feature representation to consistently achieve reliable attribution performance across all source classes.

To address these challenges, we propose a multi-view and confusion-guided ensemble framework for the Deep Learning and Mathematical Methods for Deepfake
Detection (DLMMDD) Workshop Synthetic Image Attribution Challenge~\cite{dlmmdd2026}. Our method combines multiple complementary architectures, including FFT-ConvNeXt~\cite{liu2022convnet}, DINOv2~\cite{oquab2023dinov2}, CLIP~\cite{radford2021learning}, and Xception~\cite{chollet2017xception}, to jointly capture frequency-domain artifacts, semantic representations, and forensic patterns. In addition, we employ extensive data augmentation strategies during training to improve robustness against the unknown post-processing operations introduced in the test set.

Beyond model ensembling, we further investigate the confusion behavior of the attribution system through confusion matrix analysis. We observe that certain source pairs, particularly Stable Diffusion 3 and Stable Diffusion 3.5, exhibit significantly higher mutual confusion due to their architectural and generative similarity. Motivated by this observation, we introduce a confusion-guided expert refinement strategy based on a dedicated binary classifier that is selectively activated when the ensemble model produces uncertain predictions between these two classes. We additionally apply class-adaptive confidence calibration to improve the discrimination capability for challenging classes such as Tencent Hunyuan.

Our final solution achieved 99.53\% on the public leaderboard and 99.20\% on the private leaderboard.

\section{Methodology}

\begin{figure}[ht]
      \centering
      \includegraphics[width=1.0\linewidth]{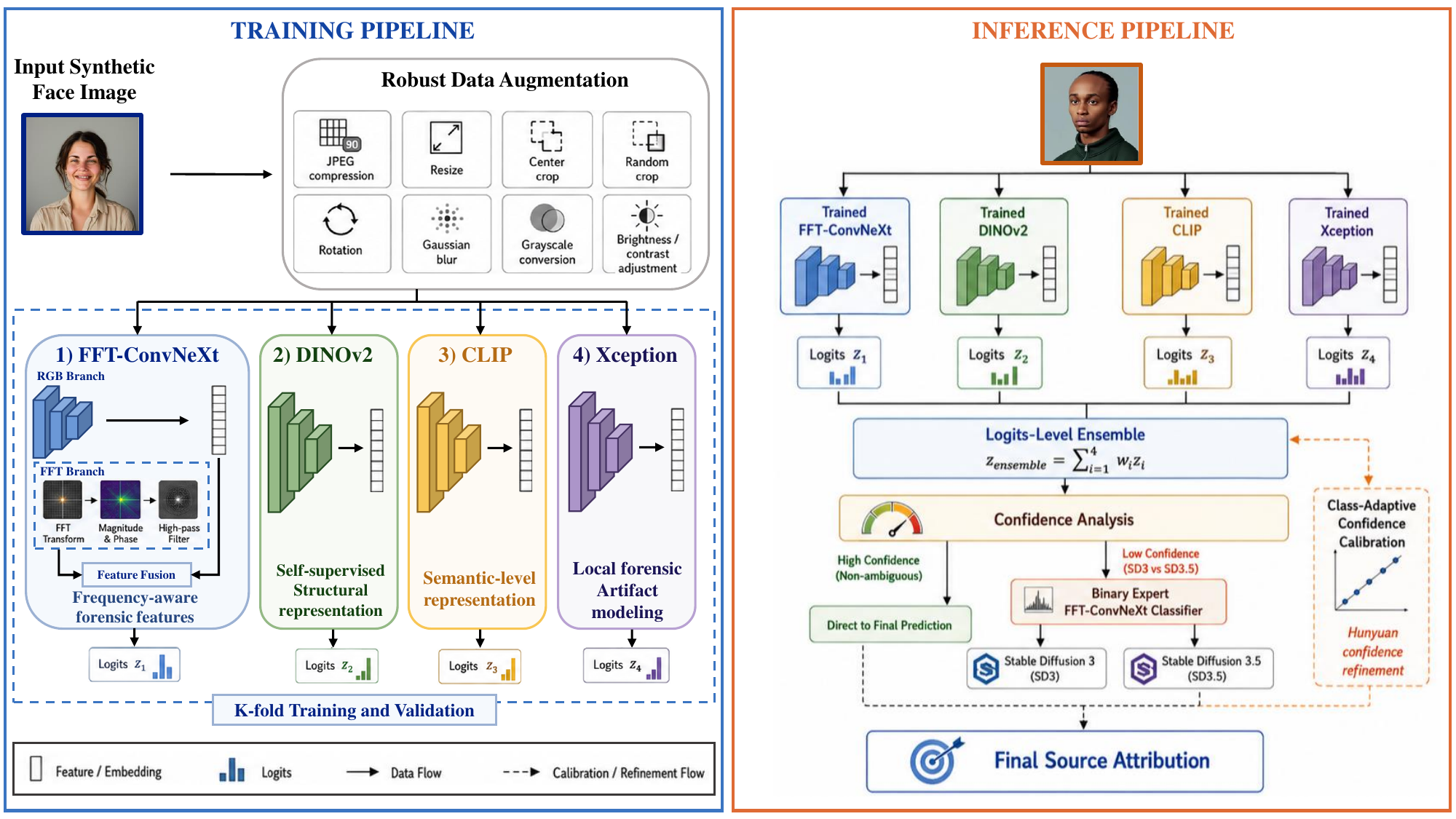}
      \caption{Overview of the proposed multi-view and confusion-guided ensemble framework. During training, synthetic face images undergo robust data augmentation and are fed into four complementary models, FFT-ConvNeXt, DINOv2, CLIP and Xception,  to learn diverse attribution features. At inference, the models generate logits that are combined via a weighted ensemble. A confusion-guided binary expert classifier selectively refines predictions, while class-adaptive confidence calibration improves predictions for challenging classes such as Tencent Hunyuan.} 
      \label{fig:pipeline}
\end{figure}

As shown in  Fig.~\ref{fig:pipeline}, our framework follows a multi-view ensemble paradigm designed for robust synthetic image attribution under diverse post-processing conditions. The overall system combines multiple complementary architectures that capture attribution signals from different perspectives, including frequency-domain artifacts, semantic representations, and forensic patterns. In addition, we introduce a confusion-guided expert refinement strategy to specifically address highly ambiguous source pairs observed during validation.

\subsection{FFT-ConvNeXt Architecture}

Among the ensemble components, we design a customized FFT-ConvNeXt model to jointly exploit spatial-domain and frequency-domain attribution cues. The motivation stems from the observation that synthetic image generators often leave subtle frequency artifacts that are difficult to capture using standard RGB representations alone, especially after image post-processing.

Given an input image $x \in \mathbb{R}^{H \times W \times 3}$, we first apply a two-dimensional Fast Fourier Transform (FFT) to obtain its frequency representation:
\begin{equation}
F(x) = \mathcal{F}(x).
\end{equation}

We then decompose the transformed spectrum into magnitude and phase components:
\begin{equation}
M = \log(1 + |F(x)|), \quad
P = \angle F(x),
\end{equation}
where logarithmic scaling is applied to stabilize the magnitude distribution.

To suppress low-frequency image content and emphasize generator-specific high-frequency traces, a high-pass mask is further applied around the spectrum center. The filtered magnitude and phase maps are concatenated to form a six-channel frequency representation:
\begin{equation}
X_{\text{fft}} = \text{Concat}(M, P).
\end{equation}

The FFT branch consists of several convolutional layers with Batch Normalization and ReLU activation for frequency feature extraction. In parallel, the RGB image is processed by a ConvNeXt-Base backbone pretrained on ImageNet. The final RGB and frequency features are concatenated and passed through a multilayer classifier:
\begin{equation}
f_{\text{fusion}} =
\text{Concat}(f_{\text{rgb}}, f_{\text{fft}}).
\end{equation}

The final prediction logits are obtained as:
\begin{equation}
z = \phi(f_{\text{fusion}}),
\end{equation}
where $\phi(\cdot)$ denotes the classification head composed of fully connected layers with dropout regularization.

The proposed FFT-ConvNeXt architecture enables the model to jointly leverage semantic image structures and frequency-domain forensic traces, improving robustness against post-processing degradations.

\subsection{Multi-View Ensemble Strategy}

Synthetic image attribution is inherently challenging because different generative models may share similar visual characteristics while differing only in subtle generation fingerprints. To improve attribution robustness and diversity, we adopt a multi-view ensemble strategy composed of four complementary architectures:

\begin{itemize}
\item \textbf{FFT-ConvNeXt}: captures frequency-domain artifacts and high-frequency forensic traces introduced by image generation pipelines.

\item \textbf{DINOv2}: provides strong self-supervised visual representations and captures global structural characteristics of generated images.

\item \textbf{CLIP}: introduces semantic-level image representations with strong robustness to image variations and distribution shifts.

\item \textbf{Xception}: serves as a forensic-oriented backbone capable of modeling subtle manipulation patterns and local image inconsistencies.
```

\end{itemize}

These models provide complementary attribution perspectives rather than relying on a single representation space. In particular, frequency-aware models are sensitive to generator fingerprints, while semantic and forensic backbones improve robustness under unknown image transformations.

During inference, the outputs of all models are combined at the logits level. Let $z_i \in \mathbb{R}^{C}$ denote the logits predicted by the $i$-th model for $C=10$ source classes. The final ensemble prediction is computed as:
\begin{equation}
z_{\text{ensemble}} =
\sum_{i=1}^{N}
w_i z_i,
\end{equation}
where $N$ denotes the number of models and $w_i$ represents the ensemble weight assigned to the $i$-th model.

The final predicted label is obtained by:
\begin{equation}
\hat{y}
=
\arg\max(z_{\text{ensemble}}).
\end{equation}

In addition, K-fold training and inference are adopted to further improve model stability and generalization performance. Predictions from different folds are averaged during inference to reduce variance and improve robustness.

\subsection{Robust Data Augmentation}

The challenge test set contains multiple unknown post-processing operations, including JPEG compression, WEBP compression, resizing, grayscale conversion, blur, rotation, and super-resolution enhancement. To improve robustness against such transformations, we employ extensive data augmentation strategies during training.

Our augmentation pipeline includes:

\begin{itemize}
\item random resizing and restoration;
\item random resized cropping;
\item center cropping;
\item horizontal flipping;
\item Gaussian blur;
\item brightness and contrast adjustment;
\item random grayscale conversion;
\item small-angle rotation;
\item random JPEG compression simulation.
\end{itemize}

These augmentations are designed to simulate realistic image degradations and distribution shifts introduced by the hidden test-time post-processing pipeline. In particular, JPEG simulation and grayscale conversion were found to substantially improve robustness against compressed and low-color-information samples. 
In addition, center cropping is randomly applied during training. Since the challenge focuses on synthetic face images, the most discriminative facial regions are typically concentrated near the image center. Random center cropping therefore helps the model focus on stable facial generation patterns while improving robustness to spatial perturbations and resizing operations. To simulate compression artifacts efficiently during training, we employ differentiable JPEG compression based on DiffJPEG\footnote{\url{https://github.com/mlomnitz/DiffJPEG}}. This strategy enables efficient and realistic compression augmentation without introducing significant computational overhead.

\subsection{Confusion-Guided Expert Refinement}

\begin{figure}[ht]
      \centering
      \includegraphics[width=1.0\linewidth]{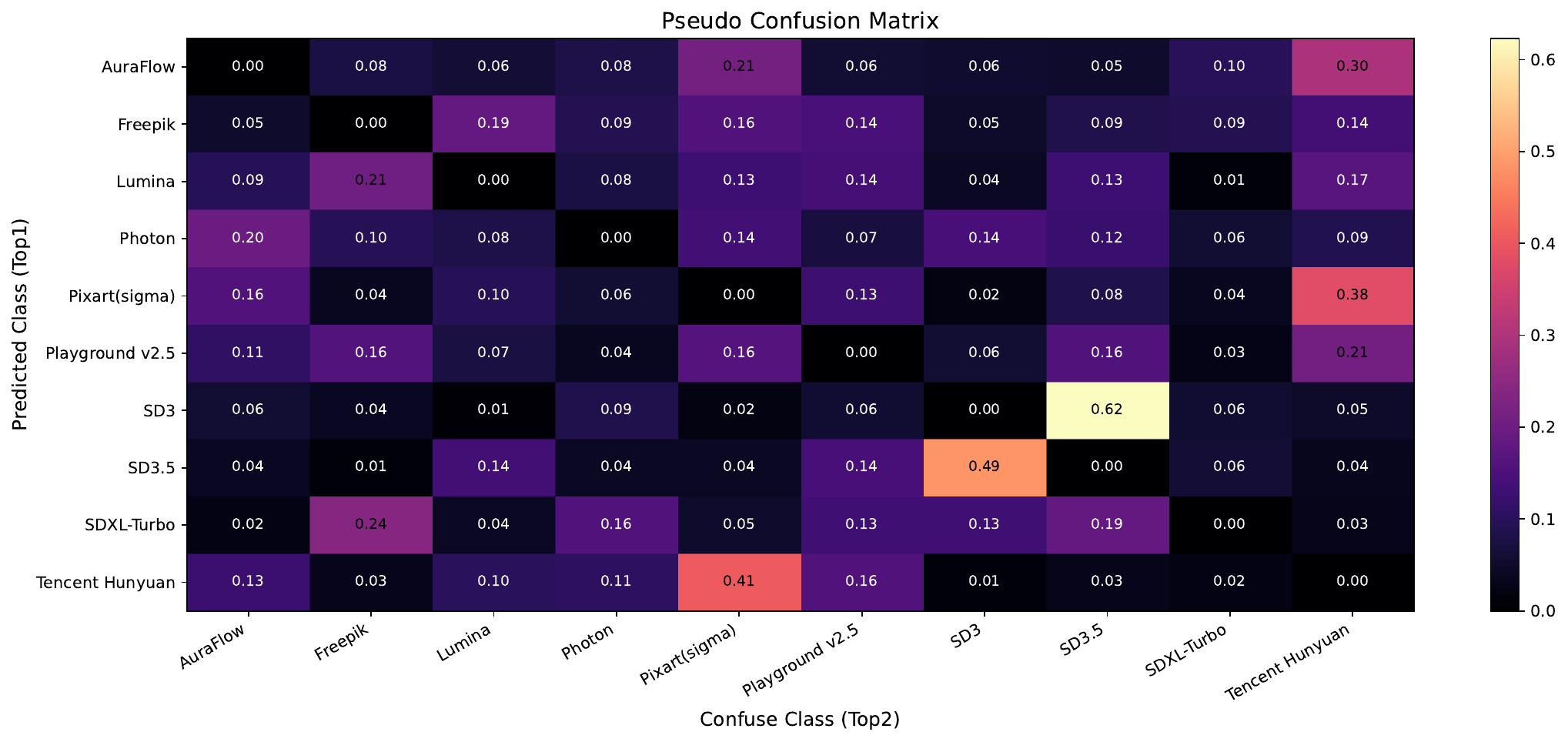}
      \caption{Normalized confusion matrix of the ensemble model on the validation set. Each row represents the predicted class (Top-1), and each column indicates the true confusing class (Top-2). Darker colors indicate higher confusion. Notably, Stable Diffusion 3 (SD3) and Stable Diffusion 3.5 (SD3.5) exhibit significant mutual confusion, motivating the design of a dedicated binary expert classifier. Tencent Hunyuan also shows partial confusion with multiple classes, which is addressed via class-adaptive confidence calibration.} \label{fig:confusion_matrix} 
\end{figure}

To better understand the failure modes of the ensemble system, we analyze the confusion matrix on the validation set. As illustrated in Fig.~\ref{fig:confusion_matrix}, we observe severe confusion between Stable Diffusion 3 and Stable Diffusion 3.5. This behavior is expected because both models belong to the same diffusion family and share highly similar generation mechanisms and visual fingerprints.

Motivated by this observation, we introduce a dedicated binary expert classifier based on the FFT-ConvNeXt architecture to specifically distinguish between these two classes. During inference, the expert model is selectively activated when the confidence difference between the two classes falls below a predefined threshold:
\begin{equation}
|p_{\text{SD3}} - p_{\text{SD3.5}}| < \tau,
\end{equation}
where $\tau$ denotes the confidence threshold.

Under this condition, the final prediction between the two candidate classes is refined using the binary expert classifier. This confusion-guided routing strategy improves discrimination performance for highly ambiguous source pairs without affecting the overall ensemble behavior.

We additionally observe that Tencent Hunyuan exhibits partial confusion with multiple classes. To alleviate this issue, we apply class-adaptive confidence calibration by slightly increasing the ensemble confidence associated with the Tencent Hunyuan category during inference. This strategy improves the absorption of uncertain samples into the correct class region and stabilizes final predictions.

\subsection{Training Details and Model Hyperparameters}

All models were initialized from publicly available pretrained weights to improve convergence and generalization performance. During training, we adopted partial fine-tuning strategies for transformer-based backbones to preserve pretrained visual representations while adapting the models to the attribution task.

We employed the AdamW optimizer with a weight decay of $1\times10^{-4}$ for all experiments. Different learning rates were assigned to backbone and classification layers to stabilize optimization. The detailed training configurations are summarized in Table~\ref{tab:training_details}.

\begin{table*}[t]
\centering
\caption{Training details and hyperparameter settings for different ensemble models.}
\label{tab:training_details}
\resizebox{\textwidth}{!}{
\begin{tabular}{l|l|l|l|l}
\hline
\textbf{Model} & \textbf{Pretrained Weights} & \textbf{Trainable Components} & \textbf{Learning Rate} & \textbf{Optimizer} \\
\hline

FFT-ConvNeXt &
ImageNet pretrained ConvNeXt &
FFT branch, classifier, full backbone &
FFT/classifier: $3\times10^{-4}$ \newline
Backbone: $1\times10^{-5}$ &
AdamW~\cite{loshchilov2017decoupled} ($wd=10^{-4}$) \\
\hline

CLIP &
OpenAI pretrained CLIP &
Classification head + last 3 vision encoder blocks &
Head: $1\times10^{-3}$ \newline
Encoder blocks: $1\times10^{-5}$ &
AdamW~\cite{loshchilov2017decoupled} ($wd=10^{-4}$) \\
\hline

DINOv2 &
Self-supervised pretrained DINOv2 &
Classification head + last 3 transformer blocks &
Head: $1\times10^{-3}$ \newline
Transformer blocks: $1\times10^{-5}$ &
AdamW~\cite{loshchilov2017decoupled} ($wd=10^{-4}$) \\
\hline

Xception &
ImageNet pretrained weights &
Classification head + full backbone &
$1\times10^{-4}$ &
AdamW~\cite{loshchilov2017decoupled} ($wd=10^{-4}$) \\
\hline

\end{tabular}
}
\end{table*}

For transformer-based architectures such as CLIP and DINOv2, only the final several encoder blocks were fine-tuned, while earlier layers remained frozen. This strategy improves training stability and reduces overfitting on the relatively limited attribution dataset. In contrast, FFT-ConvNeXt and Xception were trained with broader backbone adaptation to better capture low-level forensic artifacts and frequency-domain generation traces.

\section{Experimental Setup}

\subsection{Dataset}

Experiments were conducted on the official dataset released for the DLMMDD Workshop Synthetic Image Attribution Challenge~\cite{dlmmdd2026}. The dataset consists of synthetic face images generated by 10 open-source text-to-image models, including AuraFlow, Freepik, Lumina, Photon, PixArt($\sigma$), Playground v2.5, Stable Diffusion 3, Stable Diffusion 3.5, Stable Diffusion XL-Turbo, and Tencent Hunyuan.

The dataset was constructed using prompts derived from the WILD dataset, where each prompt was used to generate images from all source models under controlled conditions. The complete dataset contains 10,000 synthetic images with balanced class distribution, where each source contributes 1,000 images. The provided training set contains 7,000 labeled images, while the test set contains 3,000 unlabeled images with hidden ground-truth annotations.

To increase the challenge difficulty and evaluate attribution robustness, the test images were subjected to multiple undisclosed post-processing operations, including image compression, resizing, cropping, blur, grayscale conversion, and super-resolution enhancement. These transformations substantially weaken source-specific generation traces and increase inter-class ambiguity.

\subsection{Hardware Environment}

All experiments were conducted on a server running Ubuntu 18.04.6 LTS, equipped with an Intel Core i9-10940X CPU and a single NVIDIA RTX 3090 Ti GPU with 24 GB memory.

\subsection{Training Configuration}

Unless otherwise specified, all models were trained using a batch size of 16 for a maximum of 50 epochs with an early stopping strategy of 10 patience epochs. Different input resolutions were adopted according to the characteristics of each backbone architecture. Specifically, FFT-ConvNeXt was trained using input images of size $256\times256$, DINOv2 and CLIP used $224\times224$ inputs, while Xception was trained with a larger input resolution of $288\times288$ to better capture fine-grained forensic artifacts. For the Confusion-Guided Expert classifier, we adopted FFT-ConvNeXt as the backbone and set the input resolution to $512\times512$, allowing the model to capture finer forensic fingerprints that distinguish highly similar generators (i.e., Stable Diffusion 3 vs Stable Diffusion 3.5).

K-fold cross-validation was employed during training to improve generalization performance and reduce prediction variance. During inference, predictions from different folds were averaged to produce the final model outputs. In the experiments, We set $K=5$. This strategy effectively reduces variance and improves robustness, leading to more stable and reliable attribution results.

\subsection{Training Cost}

Table~\ref{tab:training_cost} summarizes the approximate computational cost and training time of the four ensemble models. The reported FLOPs correspond to the estimated training computation per image, including both forward and backward propagation.

\begin{table}[t]
\centering
\caption{Approximate training cost of different ensemble models.}
\label{tab:training_cost}
\begin{tabular}{l|c|c}
\hline
\textbf{Model} & \textbf{Training FLOPs} & \textbf{Training Time} \\
 & \textbf{(GFLOPs)} & \textbf{(s/epoch)} \\
\hline
FFT-ConvNeXt & 45.50 & 140 \\
DINOv2 & 43.94 & 99 \\
CLIP & 33.73 & 94 \\
Xception & 9.15 & 103 \\
\hline
\end{tabular}
\end{table}

FFT-ConvNeXt exhibits the highest computational complexity due to the additional frequency-domain processing branch and feature fusion operations. Although Xception has significantly lower FLOPs, its training time remains relatively high because of implementation overhead and high-resolution forensic feature extraction. Overall, the multi-view ensemble achieves a favorable balance between attribution performance and computational efficiency.

\section{Results}

\subsection{Inference Configuration}

During inference, we adopt a weighted logits-level ensemble strategy to combine predictions from the four base models, including FFT-ConvNeXt, DINOv2, CLIP, and Xception. The corresponding ensemble weights are empirically set as $[0.2, 0.4, 0.3, 0.1]$, reflecting the relative contribution of each model to the final attribution performance.

In particular, DINOv2 and CLIP are assigned higher weights due to their stronger generalization capability under distribution shifts and post-processing perturbations, while FFT-ConvNeXt and Xception contribute complementary frequency-domain and forensic-local cues.

Additionally, for the Confusion-Guided Expert classifier targeting SD3 and SD3.5, we set a trigger threshold of 0.5: the expert classifier is activated if the top-1 and top-2 predicted probabilities correspond to SD3 and SD3.5 and their difference is within 0.5. For Tencent Hunyuan, the class-adaptive confidence calibration is applied with a threshold of 0.2 to refine predictions for this partially ambiguous class. The calibration threshold was selected using the validation set and fixed before test-time inference.

\subsection{Public Leaderboard Performance}

Table~\ref{tab:leaderboard_results} reports the public leaderboard performance of individual models and the proposed ensemble framework. Among the single-model approaches, DINOv2 achieved the best standalone performance, demonstrating the effectiveness of self-supervised visual representations for synthetic image attribution. FFT-ConvNeXt also achieved strong results, indicating that frequency-domain forensic cues provide highly discriminative attribution information.

By combining multiple complementary models through logits-level ensemble learning, the attribution performance was further improved. In addition, K-fold aggregation significantly enhanced prediction stability and generalization capability, resulting in the best public leaderboard score of 0.995333. On the hidden private leaderboard, the proposed system achieved 99.20\%, indicating that the observed performance generalizes beyond the public evaluation subset.

\begin{table}[t]
\centering
\caption{Public leaderboard performance of different models and ensemble strategies.}
\label{tab:leaderboard_results}
\begin{tabular}{l|c}
\hline
\textbf{Method} & \textbf{Public Score} \\
\hline
FFT-ConvNeXt & 0.969333 \\
DINOv2 & 0.976000 \\
CLIP & 0.955333 \\
Xception & 0.958000 \\
\hline
Ensemble & 0.987333 \\
Ensemble + K-fold & \textbf{0.995333} \\
\hline
\end{tabular}
\end{table}

\subsection{Ablation Study}

We further evaluate the effectiveness of the proposed confusion-guided refinement strategies through ablation experiments.

\subsubsection{Effectiveness of the SD3/SD3.5 Expert Classifier}

As discussed in Section~3, Stable Diffusion 3 and Stable Diffusion 3.5 exhibit significant mutual confusion due to their highly similar generation characteristics. To address this issue, we introduce a dedicated binary expert classifier that is selectively activated under low-confidence conditions.

Table~\ref{tab:ablation_sd3} shows that the expert classifier improves the public leaderboard score from 0.992666 to \textbf{0.995333}, demonstrating the effectiveness of confusion-guided refinement for highly ambiguous source pairs.

\begin{table}[t]
\centering
\caption{Ablation study of the SD3/SD3.5 expert classifier.}
\label{tab:ablation_sd3}
\begin{tabular}{l|c}
\hline
\textbf{Setting} & \textbf{Public Score} \\
\hline
Without expert classifier & 0.992666 \\
With expert classifier & \textbf{0.995333} \\
\hline
\end{tabular}
\end{table}

\subsubsection{Effectiveness of Tencent Hunyuan Confidence Calibration}

We additionally evaluate the proposed class-adaptive confidence calibration strategy for Tencent Hunyuan. As observed in the confusion analysis, Tencent Hunyuan exhibits partial ambiguity with several other source classes.

As shown in Table~\ref{tab:ablation_hunyuan}, applying confidence calibration improves the final public leaderboard score from 0.993333 to \textbf{0.995333}, indicating that class-adaptive refinement helps stabilize predictions for difficult classes.

\begin{table}[t]
\centering
\caption{Ablation study of Tencent Hunyuan confidence calibration.}
\label{tab:ablation_hunyuan}
\begin{tabular}{l|c}
\hline
\textbf{Setting} & \textbf{Public Score} \\
\hline
Without calibration & 0.993333 \\
With calibration & \textbf{0.995333} \\
\hline
\end{tabular}
\end{table}

\section{Reproducibility Statement}

All results reported in this report can be fully reproduced using the provided code repository. The reproduction procedure is as follows:

\begin{enumerate}
    \item Train the four baseline classification models (FFT-ConvNeXt, DINOv2, CLIP, and Xception) with K-fold cross-validation by running:
    \begin{verbatim}
    python train.py
    \end{verbatim}
    The trained model weights are saved under the \texttt{checkpoints/} directory.

    \item Train the Confusion-Guided Expert classifier by running:
    \begin{verbatim}
    python train_bin.py
    \end{verbatim}
    The trained expert model weights are also stored in \texttt{checkpoints/}.

    \item Obtain the K-fold ensemble logits on the test set by running:
    \begin{verbatim}
    python infer.py
    \end{verbatim}
    The resulting logits are saved in the \texttt{logits\_result/} directory.

    \item Perform the final weighted ensemble, activate the Confusion-Guided Expert classifier when applicable, and apply class-adaptive confidence calibration for Tencent Hunyuan by running:
    \begin{verbatim}
    python softvote.py
    \end{verbatim}
    The final test set predictions are saved in the \texttt{results/} directory.
\end{enumerate}

This stepwise procedure, combined with the provided configuration files and data preprocessing routines, ensures full reproducibility of our results on the Synthetic Image Attribution Challenge dataset.

\section{Conclusion}

In this report, we presented a multi-view and confusion-guided ensemble framework for robust synthetic image attribution. By leveraging complementary models—FFT-ConvNeXt, DINOv2, CLIP, and Xception—at the logits level, combined with targeted confusion-guided refinement strategies, our approach effectively addresses inter-class ambiguities, particularly between highly similar generators. All code and training protocols are publicly released to ensure reproducibility, facilitating further research in synthetic image forensics and attribution.

\bibliographystyle{plain}
\bibliography{references}

\end{document}